\documentclass[10pt,twocolumn,letterpaper]{article}

\usepackage[pagenumbers]{wacv} 

\definecolor{wacvblue}{rgb}{0.21,0.49,0.74}
\usepackage[pagebackref,breaklinks,colorlinks,allcolors=wacvblue]{hyperref}

\usepackage{algorithm}
\usepackage{algorithm,algpseudocode}
\usepackage{algorithmicx}

\usepackage{graphicx}
\usepackage{amssymb}
\usepackage{booktabs}

\usepackage[utf8]{inputenc} 
\usepackage[T1]{fontenc}    
\usepackage{url}            
\usepackage{booktabs}       
\usepackage{amsfonts}       
\usepackage{nicefrac}       
\usepackage{microtype}      
\usepackage{xcolor}         

\usepackage{color}
\usepackage{amsmath}
\usepackage{amssymb}
\usepackage{mathtools}
\usepackage{arydshln}
\usepackage{amsthm}
\usepackage{bm}
\usepackage{enumitem}
\usepackage{multirow}
\usepackage{cuted}

\def\wacvPaperID{2017} 
\def\confName{WACV}
\def\confYear{2026}

\title{Point2Pose: A Generative Framework for 3D Human Pose Estimation \\ with Multi-View Point Cloud Dataset}

\author{
Hyunsoo Lee\textsuperscript{\normalfont 1}\thanks{\;indicates the corresponding author.} \hspace{10mm} 
Daeum Jeon\textsuperscript{\normalfont 2,3} \hspace{10mm} 
Hyeokjae Oh\textsuperscript{\normalfont 2,3}  \\ 
   \textsuperscript{1} ECE, Seoul National University \hspace{5mm}  \textsuperscript{2} CS, KAIST \hspace{5mm} \textsuperscript{3} Soulart Inc.\\
    \\ 
   {\tt\small philip21@snu.ac.kr, \{daumi,\,boerck\}@kaist.ac.kr}
    \vspace{-12mm}
}

\begin{document}
\maketitle

\begin{strip}
\centering
\includegraphics[width=0.9\linewidth]{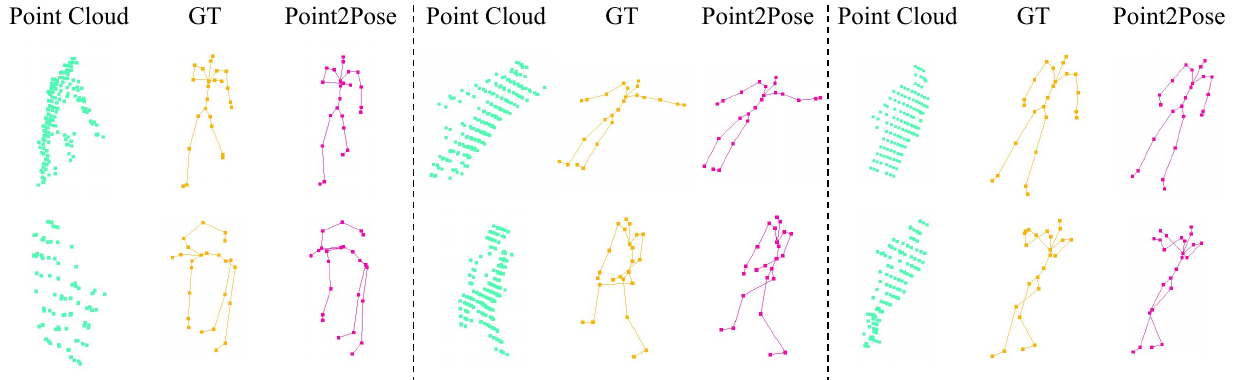} 
\captionof{figure}{
We visualize pose estimation result with Point2Pose using point clouds sampled from the LiDARHuman26M~\cite{li2022lidarcap} and  LIPD~\cite{ren2023lidar} dataset.
Here, we use ground-truth poses as input of previous frames.
Details are explained in Section~\ref{sec:exp}.
\label{fig:teaser}
\vspace{-4pt}
}
\end{strip}


\begin{abstract}
We propose a novel generative approach for 3D human pose estimation. 
3D human pose estimation poses several key challenges due to the complex geometry of the human body, self-occluding joints, and the requirement for large-scale real-world motion datasets.
To address these challenges, we introduce Point2Pose, a framework that effectively models the distribution of human poses conditioned on sequential point cloud and pose history.
Specifically, we employ a spatio-temporal point cloud encoder and a pose feature encoder to extract joint-wise features, followed by an attention-based generative regressor.
Additionally, we present a large-scale indoor dataset MVPose3D, which contains multiple modalities, including IMU data of non-trivial human motions, dense multi-view point clouds, and RGB images.
Experimental results show that the proposed method outperforms the baseline models, demonstrating its superior performance across various datasets.
\vspace{-4mm}
\end{abstract}


\section{Introduction}
\label{sec:intro}

3D Human Pose Estimation (3D HPE) is a critical research area for modeling human-computer interaction, aiming to estimate the coordinates and rotations of human joint from sensor data (\textit{e.g.}, RGB images, point clouds).
With the rise of recent research efforts, 3D HPE has been widely applied in various fields, including augmented reality~\cite{cheng2024real}, medical systems~\cite{hu2022acrnet, shin2020multi}, industry~\cite{paudel2022industrial, boldo2024real}, and autonomous driving~\cite{zanfir2023hum3dil, bauer2023weakly, zheng2022multi}.
Despite its importance, 3D HPE remains as ill-posed problem due to:
\textbf{1) The complexity of human body modeling}: 
The intricate geometry of the human body, diversity of human poses, and self-occlusions require models with strong representation power to accurately estimate diverse poses.
\textbf{2) The need for large-scale, real-world datasets}:
To effectively model the human pose data, real-world datasets with diverse human motions captured from various angles are required.
However, existing datasets often contain only trivial motions or provide sparse, single-view data, and large-scale datasets with diverse, multi-view, and dense point cloud data are scarce.

To address the issue, existing approaches often use 2D images or point clouds as input to estimate the human pose.
However, prior works that leverage 2D images~\cite{wang2024textditextpose, simoni2024depth, holmquist2023diffpose, gong2023diffpose, sengupta2021probabilistic, garau2021deca} generally show limited performance, as 2D images lack full 3D spatial information, leading to an ill-posed 2D-to-3D lifting problem.
Therefore, methods that using point clouds as input have been proposed~\cite{ren2024livehps, jang2024elmo, jang2023movin, cai2023pointhps, ren2023lidar,  li2022lidarcap, ballester2025spike}. 
Despite the advantage of spatial information, these methods still fail to capture the fine details of human pose and accurately estimate self-occluded joints, making it difficult to fully address the challenges of 3D HPE.

Recently, generative modeling~\cite{ho2020denoising, song2020denoising, lipman2022flow, tong2023improving} has shown a strong ability to model complex 3D data. 
One of the representative generative approaches, the diffusion model, has achieved state-of-the-art results in point cloud generation~\cite{ren2024tiger, nakayama2023difffacto, luo2021diffusion}, scene generation~\cite{po2024compositional, li2024sat2scene, ju2024diffindscene}, and mesh texturing~\cite{liu2024text, yu2023texture, cao2023texfusion}. 
They have also been applied to human and hand pose estimation from images~\cite{gong2023diffpose, nakayama2023difffacto, feng2023diffpose, holmquist2023diffpose, cheng2024handdiff}, highlighting their potential for pose estimation tasks.

Motivated by this, we propose \textit{Point2Pose}, a generative framework that directly estimates the 3D human pose from raw point clouds without relying on intermediate 2D representations. 
Unlike prior generative models based on 2D-to-3D lifting~\cite{gong2023diffpose, holmquist2023diffpose, feng2023diffpose} or conditional VAEs~\cite{jang2023movin}, our approach applies conditional diffusion~\cite{ho2020denoising, song2020denoising} and an optimal-transport conditional flow matching~\cite{lipman2022flow, tong2023improving} directly on 3D point clouds, which is the first application of this mechanism to point cloud–based pose estimation.
To further enhance performance, we introduce a joint-wise pose–point encoder with spatio-temporal attention that models joint–point interactions. 
This design effectively captures geometric and temporal patterns across frames, and yields substantial performance gains.

Finally,  we introduce \textit{MVPose3D}, a large-scale indoor dataset featuring dense multi-view point clouds, synchronized IMUs, and multi-view RGBs.
Unlike existing datasets, MVPose3D contains various motions and self-occlusion scenarios, offering a robust and strong benchmark for 3D HPE research.
The key contributions of our work are summarized as follows:
\begin{itemize}[label=$\bullet$]
	\item We propose Point2Pose, the first generative framework that directly estimates 3D human pose from raw point clouds via diffusion and conditional flow matching without requiring 2D-to-3D lifting. 
	\item We construct MVPose3D, a large-scale multi-modal dataset with dense multi-view point clouds, IMU, and RGB, covering diverse and complex human motions for real-world evaluation.
	\item Experimental results using various datasets demonstrate that our method achieves superior performance compared to the baselines.
\end{itemize}
%


\section{Related work}
\label{sec:related_work}

\paragraph{3D human pose estimation from point clouds.}
\label{subsec:related_work_3d_hpe}

Previous approaches in 3D human pose estimation~\cite{ren2023lidar, jang2023movin, ren2024livehps, jang2024elmo, ballester2025spike} aims to regress human joint coordinates and rotations from input point clouds and IMU data.
LiDAR-aid Inertial Poser (LIP)~\cite{ren2023lidar} utilizes GRU-based inverse kinematics solver to estimate SMPL~\cite{SMPL:2015} pose parameters and joint coordinates from sequential inputs, followed by an additional module for regressing global body orientation.
LiveHPS~\cite{ren2024livehps} predicts initial joints from consecutive point clouds using features distilled from SMPL mesh vertices, then refines them with attention mechanism while estimating full SMPL parameters.
SPiKE~\cite{ballester2025spike} partitions point clouds into volumes to extract features, then aggregates them to regress joint coordinates.

\paragraph{Generative methods for 3D HPE.}

Several methods~\cite{gong2023diffpose, feng2023diffpose, jang2023movin, jang2024elmo} leverage generative approaches for 3D human pose estimation.
DiffPose (\textit{Gong et al.})~\cite{gong2023diffpose} formulates monocular pose estimation as a reverse diffusion process~\cite{ho2020denoising, song2020denoising}, using GMM-based forward diffusion and context-conditioned reverse diffusion for 2D-to-3D lifting. 
DiffPose (\textit{Feng et al.})~\cite{feng2023diffpose} treats video pose estimation as conditional denoising diffusion, reconstructing keypoint heatmaps from corrupted Gaussian noise with a pose-Decoder. 
MOVIN~\cite{jang2023movin} employs conditional VAE~\cite{kingma2013auto} with a feature encoder and pose generator for auto-regressive pose prediction. 
ELMO~\cite{jang2024elmo} uses transformer~\cite{vaswani2017attention} to predict motion tokens corresponding to future frames conditioned on past motion history and point tokens.
Unlike previous works, our method leverages the strong generative priors of diffusion and conditional flow matching operating directly in the point cloud domain, thereby avoiding any intermediate 2D-to-3D lifting.

\paragraph{Real-world human motion datasets for 3D HPE.}
\label{subsec:related_work_datasets}

Existing human motion capture datasets~\cite{vonMarcard2018, h36m_pami, li2022lidarcap, jang2023movin, ren2024livehps} provide diverse modalities for 3D HPE.
For example, 3DPW~\cite{vonMarcard2018} and Human3.6M~\cite{h36m_pami} datasets consist of large-scale human motion data with RGB images. 
However, these datasets lack 3D point clouds, which limits the usability of each dataset for 3D input modalities.
Datasets with point clouds often utilize LiDAR sensors for data acquisition.
LiDARHuman26M~\cite{li2022lidarcap} captures outdoor motion with a single LiDAR.
MOVIN dataset~\cite{jang2023movin} collects indoor motion with a single LiDAR and IMUs, resulting in both static and locomotive motions.
FreeMotion~\cite{ren2024livehps} acquires 40 motion types with multi-view LiDARs and SMPL~\cite{SMPL:2015} parameters in a panoptic studio.
While multiple existing datasets often cover simple motions, with sparse, single-view data, our dataset offers large-scale data with diverse, dense, multi-view-taken point clouds.


\section{Proposed method}
\label{sec:method}

\begin{figure*}[t!]
	\centering
	\includegraphics[width=0.95\linewidth]{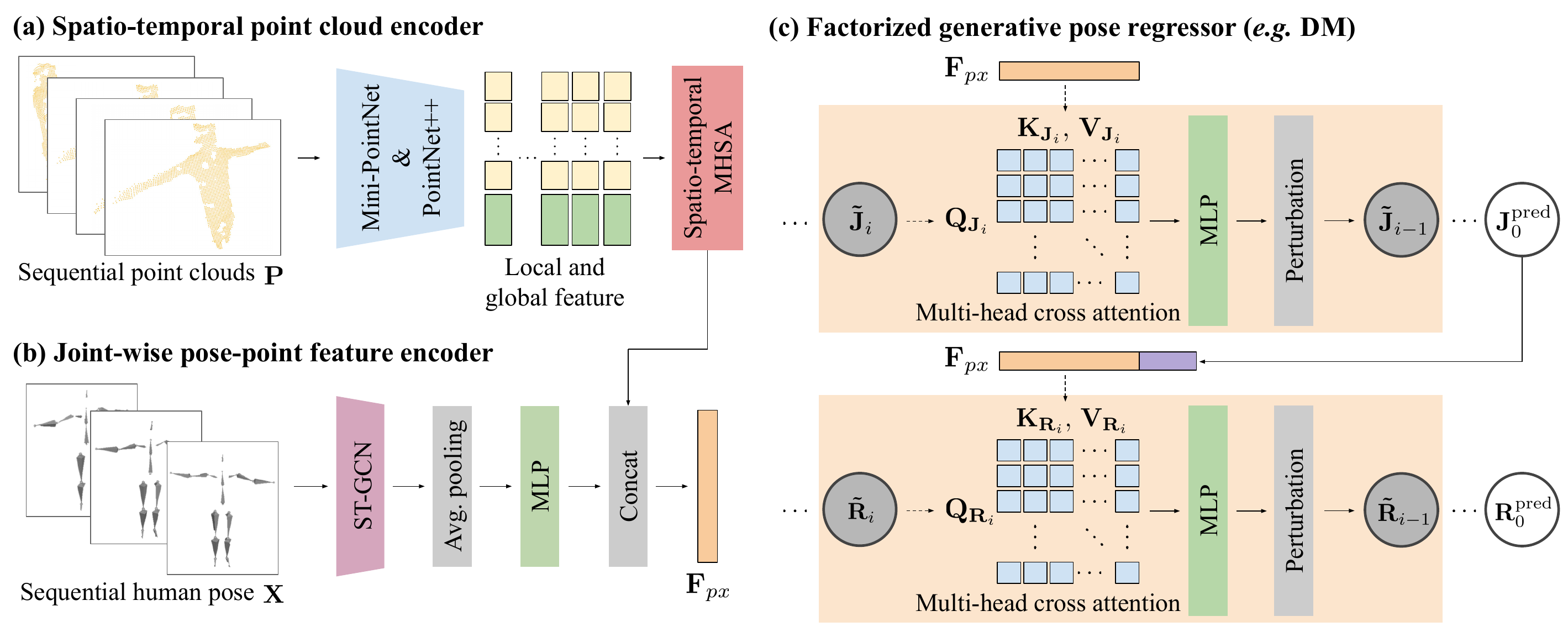}
    \vspace{-3mm}
	\caption{Overview of the proposed method for 3D human pose estimation.
    We visualize the case of utilizing diffusion models (DM).}
	\vspace{-4mm}
\label{fig:method}
\end{figure*}

\subsection{Model overview}
\label{subsec:method_preliminaries}
We introduce Point2Pose, a novel generative framework for 3D human pose estimation (HPE). 
The input to our model consists of two variables, sequential point clouds $\mathbf{P} \in \mathbb{R}^{T \times N \times 3}$ and sequential pose history $\mathbf{X} \in \mathbb{R}^{(T-1) \times J \times 9}$, where $T$ is the number of continuous frames, $N$ denotes the number of points in the point cloud of a single frame, and $J$ corresponds to the number of joints.
Note that pose of a single frame $\mathbf{X}(t) \in \mathbb{R}^{J \times 9}$ is composed of relative root coordinates $\mathbf{J}(t) \in \mathbb{R}^{J \times 3}$ and rotations $\mathbf{R}(t) \in \mathbb{R}^{J \times 6}$.
We use 6D representations~\cite{zhou2019continuity} for joint rotations.
Then, our model predicts the pose data of the current frame.

Point2Pose is composed of three main components.
Spatio-temporal point cloud encoder (Sec.~\ref{subsec:method_pc_enc}) encodes the input point clouds by extracting local and global features. 
Joint-wise pose-point feature encoder (Sec.~\ref{subsec:method_imu_enc})  then encodes the pose history by utilizing cross-attention with extracted point cloud feature.
Lastly, joint coordinates and rotations are predicted with Factorized generative pose regressor (Sec.~\ref{subsec:method_imu_reg}).
Figure~\ref{fig:method} summarizes the proposed method.

\subsection{Spatio-temporal point cloud encoder}
\label{subsec:method_pc_enc}

\subsubsection{Point cloud feature extraction}

We introduce patch-wise point cloud features by leveraging both local and global information to effectively model the complex geometry of the human body. 
Comprehensive knowledge on the human body structure is represented by global features, while local patches provide detailed information on each part of the human body.
Following Point-BERT~\cite{yu2022point}, we use farthest point sampling (FPS)~\cite{eldar1997farthest} to sample $G$ patch centers from the input point cloud $\mathbf{P}$, and apply k-nearest neighbors (kNN) algorithm to select $M-1$ nearest points for each center, forming local patches $\mathbf{\bar{P}} \in \mathbb{R}^{T \times G \times M \times 3}$. 
Mini-PointNet~\cite{qi2017pointnet} is applied on each local patch to extract local features  $\mathbf{F}_{\mathrm{local}} \in \mathbb{R}^{T \times G \times k_1}$, while global features $\mathbf{F}_{\mathrm{global}} \in \mathbb{R}^{T \times k_2}$ are extracted using PointNet++~\cite{qi2017pointnet++}. 
These are concatenated to construct a novel patch-wise point cloud features $\mathbf{F}_p \in \mathbb{R}^{T \times G \times k_3}$:
\begin{equation}
    \mathbf{F}_p = [\mathbf{F}_{\mathrm{local}}; \mathrm{Duplicate}(\mathbf{F}_{\mathrm{global}}, G)],
    \label{eq:pc_feature}
\end{equation}
where $\mathrm{Duplicate} (\cdot, k)$ operator duplicates the input variable by $k$ times, $[\cdot \; ; \cdot]$ denotes concatenation, and $k_3=k_1 + k_2$.

\subsubsection{Spatio-temporal MHSA module}
To obtain a powerful representation of the point cloud, we effectively fuse spatio-temporal information using attention mechanism~\cite{vaswani2017attention}, which is given by
\begin{equation}
    \mathrm{Attn}(Q , K, V) = \mathrm{Softmax} \left( \frac{QK^{\top}}{\sqrt{d_k}}\right)V,
    \label{eq:attention}
\end{equation}
where $d_k$ is the dimension of $Q$ and $K$.
Modeling human body geometry from sequential point clouds is analogous to spatio-temporal representation learning in video sequences~\cite{arnab2021vivit,bertasius2021space, fan2021multiscale, ranasinghe2022self}.
Therefore it's plausible to adapt spatio-temporal attention mechanism.
Motivated by the Video Vision Transformer (ViViT)~\cite{arnab2021vivit}, we employed factorized spatio-temporal transformer to utilize multi-head self-attention (MHSA)~\cite{vaswani2017attention}.
Then, the spatio-temporal point cloud feature $\mathbf{F}_p^{st} \in \mathbb{R}^{k_p}$ is calculated by
\begin{align}
    \mathbf{F}_p^{t} &= f^{\mathrm{t}}(\mathbf{F}_p) = \mathrm{Attn}(f^Q_t(\mathbf{F}_p), f^K_t(\mathbf{F}_p), f^K_t(\mathbf{F}_p)), \nonumber \\
    \mathbf{F}_p^{st} &= f^{\mathrm{s}}(\mathbf{F}_p^t) = \mathrm{Attn}(f^Q_s(\mathbf{F}_p^t), f^K_s(\mathbf{F}_p^t), f^K_s(\mathbf{F}_p^t)), 
    \label{eq:spatio_temporal_attn}
\end{align}
where $f^{\mathrm{t}}(\cdot)$ and $f^{\mathrm{s}}(\cdot)$ denotes transformer-based MHSA module for temporal and spatial branch, respectively.
Both modules generate $Q$, $K$ using the projection $f^Q(\cdot)$, $f^K(\cdot)$ of the input features.
For the simplicity, we omit the notation of positional encoding implemented in the transformer network for the rest of the paper.

\subsection{Joint-wise pose-point feature encoder}
\label{subsec:method_imu_enc}
To model the prior knowledge of past movements of human in predicting current human pose, we encode pose history $\mathbf{X}$ from the previous $T-1$ frames using Spatio-Temporal Graph Convolutional Networks (ST-GCN)~\cite{yan2018spatial}, which captures graphical information over consecutive frames of the human body skeleton.
The resulting joint-wise feature $\mathbf{F}_x \in \mathbb{R}^{J \times k_x}$ is computed as:
\begin{equation}
    \mathbf{F}_x = f^\mathrm{MLP}(\mathrm{AvgPool}(f^{\mathrm{ST-GCN}}(\mathbf{X}))),
    \label{eq:imu_feature}
\end{equation}
where $\mathrm{AvgPool}(\cdot)$ fuses temporal components of ST-GCN output using averaging pooling. 
We use additional fully-connected network $f^{\mathrm{MLP}}$ to strengthen the representation power of $\mathbf{F}_x$. Furthermore, to aggregate pose and point cloud feature, we concat $\mathbf{F}_x$ and $\mathbf{F}_p^{st}$, resulting in joint-wise pose-point feature $\mathbf{F}_{px} \in \mathbb{R}^{J \times k_{px}}$ as follows:
\begin{equation}
    \mathbf{F}_{px} = [\mathbf{F}_{x}; \mathrm{Duplicate}(\mathbf{F}_{p}, J)],
    \label{eq:pc_imu_feature}
\end{equation}
where $k_{px}=k_p + k_x$.

\subsection{Factorized generative pose regressor}
\label{subsec:method_imu_reg}

To estimate the distribution of joint coordinates and rotations conditioned on complex human body structure, we employ a generative approach~\cite{ho2020denoising, song2020denoising, lipman2022flow, tong2023improving} to predict pose data $\mathbf{X}^{\mathrm{pred}} = [\mathbf{J}^{\mathrm{pred}}; \mathbf{R}^{\mathrm{pred}}]$. 
We perturb the ground truth to obtain noisy samples $\mathbf{\tilde{X}}_i = [\mathbf{\tilde{J}}_i; \mathbf{\tilde{R}}_i]$ using diffusion models (\textbf{DM}) and optimal-transport conditional flow matching (\textbf{OT-CFM}), where $i \in [0, I]$ is the denoising iteration used for sampling.
We then denoise these samples to predict $\mathbf{X}_i^{\mathrm{pred}}$ through a factorized branch:
\begin{align}
    \mathbf{{J}}^{\mathrm{pred}}_i &= f^\mathrm{Denoiser}_{\mathbf{J}}(\mathbf{\tilde{J}}_i, \mathbf{F}_{px}, e_i, e_{J}), \nonumber \\
    \mathbf{{R}}^{\mathrm{pred}}_i &= f^\mathrm{Denoiser}_{\mathbf{R}}(\mathbf{\tilde{R}}_i, \mathbf{{J}}^{\mathrm{pred}}_i,\mathbf{F}_{px}, e_i, e_{J}),
    \label{eq:denoising_coord_rotation} \\
    \mathbf{{X}}^{\mathrm{pred}}_i &:= [\mathbf{{J}}^{\mathrm{pred}}_i ; \mathbf{{R}}^{\mathrm{pred}}_i],
    \label{eq:denoising_overview}
\end{align}
where $e_i$ and $e_J$ are denoising iteration and joint index embeddings.
The factorized design promotes better kinematic consistency by first predicting joint coordinates which provides a strong prior for estimating rotations.
Then, an auxiliary rotation prediction branch further guides $f^\mathrm{Denoiser}_{\mathbf{R}}$ to learn kinematic dependencies.

Unlike prior works~\cite{li2022lidarcap, jang2023movin, jang2024elmo, gong2023diffpose, holmquist2023diffpose, feng2023diffpose}, which utilize PointNet++~\cite{qi2017pointnet, qi2017pointnet++}, GCN~\cite{kipf2016semi, yan2018spatial}, or generative processes~\cite{ho2020denoising, song2020denoising}, our method
1) factorizes the complex geometry of human joints by leveraging both local and global features, 
2) leverages joint-wise pose-point feature that effectively extracts consecutive geometry patterns with spatio-temporal attention, and 
3) estimates pose data via a novel conditional generative approach. 
We further demonstrate the effectiveness of each proposed component in Sec~\ref{subsec:exp_ablation}.
Following sections discuss noise perturbation and denoising procedure.

\subsubsection{Noise perturbation mechanism}
\label{subsubsec:method_noise_perturbation}
We denote each model utilizing DM and OT-CFM as Point2Pose$_{\mathrm{DM}}$ and Point2Pose$_{\mathrm{OT-CFM}}$, respectively.
Since we utilize factorized denoising branch between $\mathbf{\tilde{J}}_i$ and $\mathbf{\tilde{R}}_i$, we also conduct noise perturbation in a factorized manner.

\noindent\textbf{DM.}
At each denoising iteration $i$, we perturb the ground truth pose data $\mathbf{X}^{\mathrm{gt}}$ to calculate $\mathbf{\tilde{X}}_i$ by following DDPM~\cite{ho2020denoising}:
\begin{align}
    \mathbf{\tilde{J}}_i = \sqrt{\bar{\alpha}_i} \mathbf{J}^{\mathrm{gt}} + \sqrt{1-\bar{\alpha}_i} \mathbf{\epsilon}, &\quad
    \mathbf{\tilde{R}}_i = \sqrt{\bar{\alpha}_i} \mathbf{R}^{\mathrm{gt}} + \sqrt{1-\bar{\alpha}_i} \mathbf{\epsilon}, \nonumber \\
    \mathbf{\tilde{X}}_i = \mathrm{Perturb}_{\mathrm{DM}}&(\mathbf{X}^{\mathrm{gt}}, i) := [\mathbf{\tilde{J}}_i; \mathbf{\tilde{R}}_i]
    \label{eq:diffusion_perturb}
\end{align}
where $\bar{\alpha}_i = \prod_{j=0}^i \alpha_j$, $\alpha_i = 1-\beta_i$, $\{\beta_i\}_{i=0}^{I}$ is decreasing sequence which controls the variance of the noise at each denoising iteration $i$, and $\mathbf{\epsilon}$ is noise sampled from $\mathcal{N}(\mathbf{0}, \mathbf{I})$.

\noindent\textbf{OT-CFM.}
We derive $\mathbf{\tilde{X}}_i$ as follows:
\begin{align}
    \mathbf{\tilde{J}}_i &= (i/I)\mathbf{Z}^{j}_I + (1 - i/I)\mathbf{Z}^j_0 + \sigma \mathbf{\epsilon} \nonumber, \\
    \mathbf{\tilde{R}}_i &= (i/I)\mathbf{Z}^r_I + (1 - i/I)\mathbf{Z}^r_0 + \sigma \mathbf{\epsilon}, \nonumber \\
    \mathbf{\tilde{X}}_i &= \mathrm{Perturb}_{\mathrm{OT-CFM}}(\mathbf{X}^{\mathrm{gt}}, i) :=[\mathbf{\tilde{J}}_i; \mathbf{\tilde{R}}_i]
\end{align}
where $\mathbf{\epsilon} \sim \mathcal{N}(0, \mathbf{I})$, $\sigma$ is the fixed variance, and ($\mathbf{Z}^j_0, \mathbf{Z}^j_I)$ is sampled from distribution $q((\mathbf{Z}^j_0, \mathbf{Z}^j_I)) = \pi (\mathbf{\epsilon}, \mathbf{J}^{\mathrm{gt}})$. 
Note that $\pi(\cdot, \cdot)$ is 2-Wasserstein optimal transport map~\cite{tong2023improving}.
Similarly, $(\mathbf{Z}^r_0, \mathbf{Z}^r_I)$ is  sampled from $q((\mathbf{Z}^r_0, \mathbf{Z}^r_I)) = \pi (\mathbf{\epsilon}, \mathbf{R}^{\mathrm{gt}})$.

\noindent\textbf{Denoising iteration and joint index embeddings.} 
\label{subsubsec:method_embeddings}
To incorporate noise perturbation level as condition for denoising, we follow DDIM~\cite{song2020denoising} to generate denoising iteration embedding $e_i \in \mathbb{R}^{J \times 3}$.
In addition, we incorporate joint index embedding $e_{J} \in \mathbb{R}^{J \times 3}$ as joint indicator vector, motivated by the effectiveness of $e_{J}$ in recent works~\cite{gong2023diffpose, cheng2024handdiff}. 
We use sinusoidal embeddings for both as follows:
\begin{align}
    e_i &= \mathrm{Duplicate}([i; \sin(i); \cos(i)], J), \nonumber\\
    e_{J} &= [j; \sin(j); \cos(j)]_{j=0}^{J-1}.
    \label{eq:embedding}
\end{align}

\subsubsection{Denoising mechanism}
\label{subsubsec:method_denoising}
We train the denoising network to predict ground truth pose data $\mathbf{J}^{\mathrm{gt}}$, $\mathbf{R}^{\mathrm{gt}}$ from the perturbed inputs $\mathbf{\tilde{J}}_i$ and $\mathbf{\tilde{R}}_i$ .
To effectively fuse the information between perturbed variables and pose-point feature $\mathbf{F}_{px}$, we use multi-head cross-attention (MHCA) mechanism~\cite{vaswani2017attention} following Eq.~\eqref{eq:attention}.
Then we rewrite Eq.~\eqref{eq:denoising_coord_rotation} by incorporating MHCA as follows:
\begin{align}
    \mathbf{{J}}^{\mathrm{pred}}_i &=f^\mathrm{Denoiser}_{\mathbf{J}}(\mathbf{\tilde{J}}_i, \mathbf{F}_{px}, e_i, e_{J}) \nonumber \\
    &:= f^W_{\mathbf{J}}(\mathrm{Attn}(Q_{\mathbf{J}_i}, K_{\mathbf{J}_i}, V_{\mathbf{J}_i})),
    \label{eq:coordinate_attn}
\end{align}
where 
\begin{equation*}
    Q_{\mathbf{J}_i} = f^Q_{\mathbf{J}}([\mathbf{\tilde{J}}_i; e_i ; e_J]), \; K_{\mathbf{J}_i}=V_{\mathbf{J}_i}=f^K_{\mathbf{J}}(\mathbf{F}_{px}),
    \label{eq:coordinate_attn_qkv}
\end{equation*}
and $f^Q_{\mathbf{J}}(\cdot)$, $f^Q_{\mathbf{J}}(\cdot)$, $f^W_{\mathbf{J}}(\cdot)$ denotes projections of input variables, followed by fully-connected layers.
We similarly derive the behavior of $f^\mathrm{Denoiser}_{\mathbf{R}}$ by
\begin{align}
    \mathbf{{R}}^{\mathrm{pred}}_i &= f^\mathrm{Denoiser}_{\mathbf{R}}(\mathbf{\tilde{R}}_i, \mathbf{{J}}^{\mathrm{pred}}_i,\mathbf{F}_{px}, e_i, e_{J}) \nonumber  \\&:= \mathbf{\tilde{R}}_i + f^W_{\mathbf{R}}(\mathrm{Attn}(Q_{\mathbf{R}_i}, K_{\mathbf{R}_i}, V_{\mathbf{R}_i})),
    \label{eq:rotation_attn}
\end{align}
where
\begin{equation*}
    Q_{\mathbf{R}_i} = f^Q_{\mathbf{R}}([\mathbf{\tilde{R}}_i; e_i ; e_J]), \; K_{\mathbf{R}_i}=V_{\mathbf{R}_i}=f^K_{\mathbf{R}}([\mathbf{{J}}^{\mathrm{pred}}_i; \mathbf{F}_{px}]).
    \label{eq:rotation_attn_qkv}
\end{equation*}
Note that we empirically induce residual connection in Eq.~\eqref{eq:rotation_attn} for better performance.

\subsection{Training and inference}
\label{subsec:method_train_inference}

\subsubsection{Training}
To train Point2Pose, we uniformly sample a denoising iteration $i$ from the range $[0, I]$.
Then we optimize the model parameters $\theta$ to accurately regress the ground-truth pose data $\mathbf{X}^{\mathrm{gt}}$ by denoising the perturbed data $\mathbf{\tilde{X}}_i$.
We use smooth L1 function which is widely adapted in the previous pose estimation methods~\cite{cheng2023handr2n2, tian2023single, gong2023diffpose}, which is given by
\begin{equation}
    \mathrm{SmoothL1}(\mathbf{x}, \mathbf{y}) = \begin{cases}
			50d^2, & \text{if $d<\tau$}\\
            d - 0.005, & \text{otherwise}
		 \end{cases}
    \label{eq:smooth_l1}
\end{equation}
where $d = |\mathbf{x} - \mathbf{y}|$ and $\tau$ is threshold value. 
The training objective for the proposed method calculated as
\begin{equation}
    \mathcal{L}_{\mathrm{}} = \frac{1}{J} \cdot \mathrm{SmoothL1}( \mathbf{{X}}^{\mathrm{pred}}_i, \mathbf{X}^{\mathrm{gt}}).
    \label{eq:loss_fn}
\end{equation}

\subsubsection{Inference}
To sample joint coordinates and rotations using the proposed method, we iteratively apply the denoising process to the perturbed pose data, starting from the initial noise sampled from  $\mathcal{N}(0, \mathbf{I})$.
As a result, a sequence $\{ \mathbf{{X}}_i^{\mathrm{pred}} \}_{i=I}^0$ is generated by progressively applying the denoising process. 
We provide a detailed explanation of the sampling strategies for the two generative approaches that we have utilized.

\noindent\textbf{DM.} 
At $i$'th iteration, we denoise the perturbed data $\mathbf{\hat{X}_i}$ into $\mathbf{{X}}^{\mathrm{pred}}_i$ using Eq.~\eqref{eq:denoising_overview}.
Then we calculate $\mathbf{\tilde{X}}_{i-1}$, using the perturbation strategy used in DDIM~\cite{song2020denoising} as follows:
\begin{align}
    \mathbf{\tilde{J}}_{i-1} &= \sqrt{\bar{\alpha}_{i-1}} \mathbf{{J}}^{\mathrm{pred}}_i + \sqrt{1 - \bar{\alpha}_{i-1} - \sigma_t^2} \mathbf{{\epsilon}}^j_i + \sigma_i \epsilon, 
    \label{eq:dm_inference_perturb_coord} \\
    \mathbf{\tilde{R}}_{i-1} &= \sqrt{\bar{\alpha}_{i-1}} \mathbf{{R}}^{\mathrm{pred}}_i + \sqrt{1 - \bar{\alpha}_{i-1} - \sigma_t^2} \mathbf{{\epsilon}}^r_i + \sigma_i \epsilon, \label{eq:dm_inference_perturb_rotation} \\
    \mathbf{\tilde{X}}_{i-1} &= [\mathbf{\tilde{J}}_{i-1}; \mathbf{\tilde{R}}_{i-1}]
    \label{eq:dm_inference_perturb}
\end{align}
where $\sigma_t = \sqrt{{(1-\bar{\alpha}_{i-1})(\bar{\alpha}_{i-1} - \bar{\alpha}_i  )}/{(\bar{\alpha}_{i-1}(1-\bar{\alpha}_i))}}$,  $\mathbf{{\epsilon}}^j_i = (\mathbf{\tilde{J}_i} - \sqrt{\bar{\alpha_i}} \mathbf{{J}}^{\mathrm{pred}}_i )/{\sqrt{1-\bar{\alpha}_i}}$, and $\epsilon \sim \mathcal{N}(0, \mathbf{I})$.
Note that $\mathbf{\epsilon}^r_i$ is calculated with $\mathbf{\tilde{R}}_i$ and $\mathbf{R}_i^{\mathrm{pred}}$.
We gradually decrease $i$ from $I$ to $0$ with the initial noise variable $\mathbf{\tilde{X}}_I \sim \mathcal{N}(0, \mathbf{I})$.
Then we take $\mathbf{{X}}_0^{\mathrm{pred}}$ as the output of our model.

\noindent\textbf{OT-CFM.}
At $i$'th iteration, we calculate $\mathbf{{X}}_i^{\mathrm{pred}}$ using the same procedure as with DM.
Then we get $\mathbf{\tilde{X}}_{i+1}$ as follows:
\begin{equation}
    \mathbf{\tilde{X}}_{i+1} = \mathrm{Perturb}_{\mathrm{OT-CFM}}(\mathbf{{X}}_i^{\mathrm{pred}}, i+1).
    \label{eq:cfm_inference_perturb}
\end{equation}
Starting from $i=0$ with $\mathbf{\tilde{X}}_0 \sim \mathcal{N}(0, \mathbf{I})$, the model outputs $\mathbf{{X}}_I^{\mathrm{pred}}$ by iteratively increasing $i$ to $I$.
The pseudo-code for the training and inference procedures of Point2Pose$_{\mathrm{DM}}$ is provided in Appendix~\ref{sec:supp_pseudo_code}.


\section{Proposed dataset}
\label{sec:dataset}

\begin{figure}[t]
    \centering
    \includegraphics[width=0.75\linewidth]{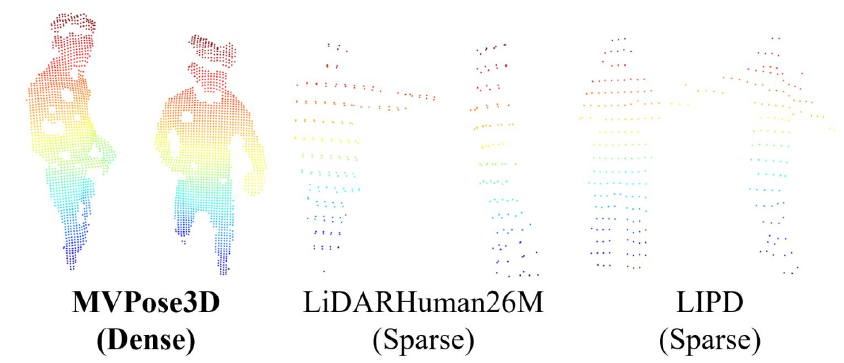}
    \vspace{-1mm}
    \caption{MVPose3D features a denser distribution than LiDARHuman26M~\cite{li2022lidarcap} and LIPD~\cite{ren2023lidar}, enhancing 3D pose estimation and motion analysis.}
    \label{fig:dataset_cmp}
    \vspace{-3mm}
\end{figure}

\begin{figure}[t]
    \centering
    \includegraphics[width=\linewidth]{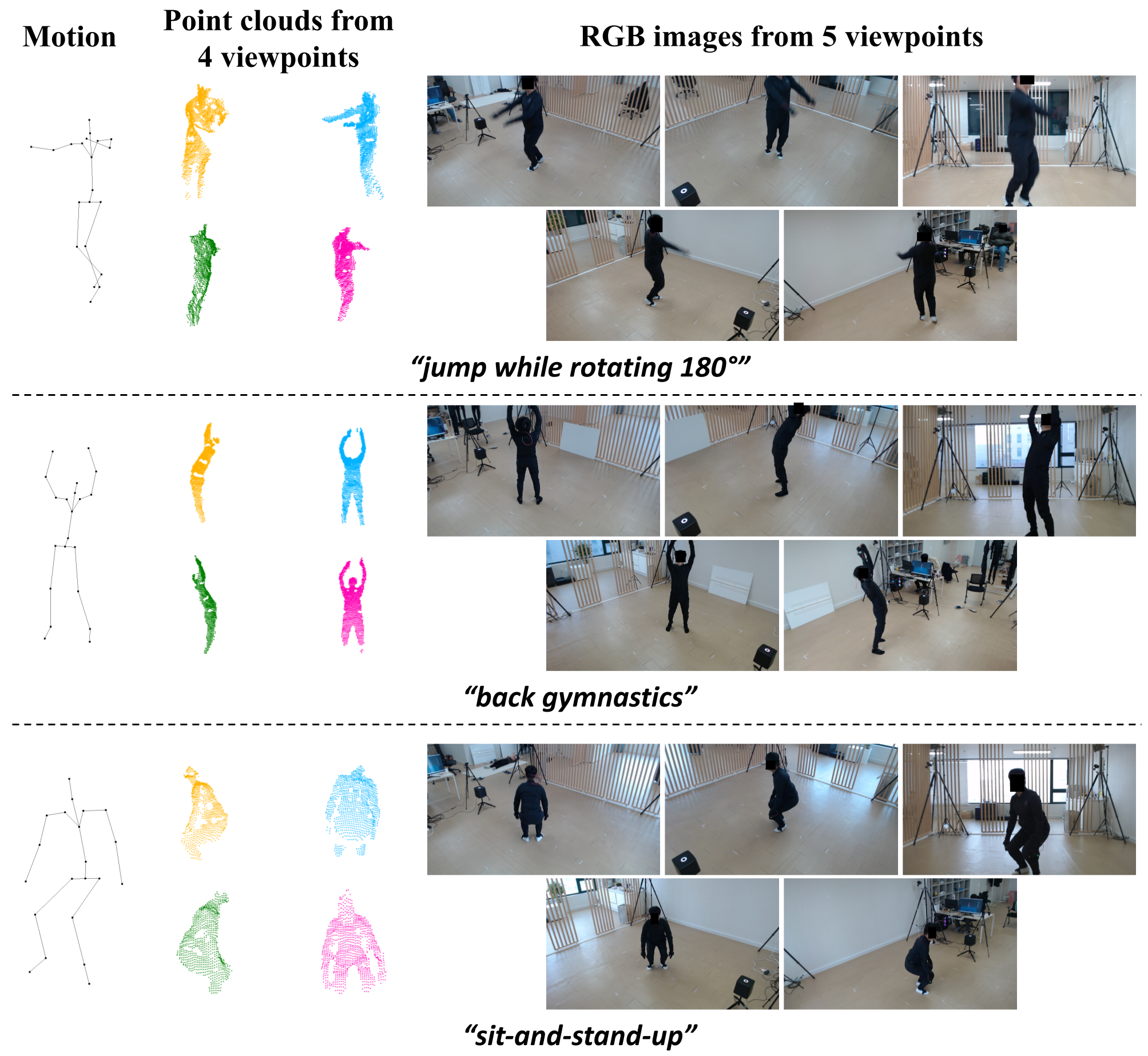}
    \caption{Samples from the proposed dataset. 
    The MVPose3D dataset includes a diverse range of human motions. 
    Each sample consists of four point clouds, five RGB images, 3D joint coordinates, and joint rotation data. 
    We mask each subject’s face for visualization.
    Best viewed when magnified.}
    \label{fig:dataset_sequence}
    \vspace{-5mm}
\end{figure}

\begin{table*}[h!]
\centering
\caption{{Comparison between various datasets constructed for 3D human pose estimation.} Our dataset provides multi-view, large-scale, and diverse modalities compared to prior datasets.}
\vspace{-2mm}
\scalebox{0.95}{
\begin{tabular}{lccccccccc}
\toprule
Name & \# of frame & \# of sbj & \# of actions & \# of view & In/Outdoor & Point cloud & RGB image & IMU \\
\midrule
Human3.6M~\cite{h36m_pami} & 3.6M & 11 & 17 & 4 & Indoor & X & O & O \\
3DPW~\cite{vonMarcard2018} & 51k & 7 & - & 1 & Outdoor & X & O & O \\
SURREAL~\cite{varol17_surreal} & 6M & 145 & 23 & 1 & Indoor & X & O & O \\
ITOP~\cite{haque2016towards} & 101k & 20 & 15 & 2 & - & O & X & X \\
LidarHuman26M~\cite{li2022lidarcap} & 184k & 13 & 20 & 1 & Outdoor & O & O & O \\
LIPD~\cite{ren2023lidar} & 62k & 15 & 30 & 1 & Outdoor & O & O & O \\
MOVIN~\cite{jang2023movin} & 161k & 10 & 29 & 1 & Indoor & O & O & O \\
SLOPER4D~\cite{Dai_2023_CVPR} & 100k & 12 & - & 1 & Outdoor & O & O & O \\
CIMI4D~\cite{yan2023cimi4d} & 180k & 12 & - & 1 & Outdoor & O & O & O \\
\hdashline
\textbf{MVPose3D (Ours)} & 215k & 12 & 18 ($\times$4) & 4-5 & Indoor & O & O & O \\
\bottomrule
\end{tabular}}
\label{tab:dataset_cmp}
\vspace{-3mm}
\end{table*}

\subsection{Acquisition}

We introduce \textit{MVPose3D}, a benchmark dataset for accurate 3D human pose estimation and motion analysis.
Our capture system includes controlled indoor setup with 5 RGB cameras, 4 indirect time-of-flight (iToF) sensors, and a motion capture suit.
The RGB cameras ($1280 \times 720$ resolution) and iToF sensors ($320 \times 240$ resolution after binning) are arranged with four RGB-iToF pairs mounted at ceiling corners and one additional RGB camera is installed on the front side. 
This system provides synchronized multi-view RGB, dense depth-based point clouds, and accurate 3D joint annotations at 30 fps.
Our dataset consists of 215,039 synchronized frames collected from 12 subjects with varying body shapes and heights (158–181 cm).
Each subject performs 18 diverse actions, repeated four times from multiple orientations to capture viewpoint variation.
The actions include both static and dynamic movements, such as gymnastics, rotating, and running, ensuring broad motion coverage.
All participants provided informed consent for the acquisition and use of their motion data in this work.
Technical details on studio dimensions, sensor properties, calibration procedures, and the full action list are provided in the Appendix~\ref{sec:supp_dataset}.

\subsection{Human body segmentation from point clouds} 
Each captured depth map from the iToF sensor is converted into a segmented point cloud representing the human body.
Background pixels are removed by subtracting a static reference depth map. 
Then, connected-component analysis using DFS~\cite{tarjan1972depth} with 8-connected neighbors retains the largest component as the human body. 
The point cloud is further denoised using DBSCAN~\cite{ester1996density} to remove outliers, followed by Statistical Outlier Removal (SOR)~\cite{rusu2010semantic} for refinement.

\subsection{Analysis}

Table~\ref{tab:dataset_cmp} compares key parameters across datasets.
Among point cloud–based datasets, MVPose3D contains more frames and diverse viewpoints than existing ones. 
While SURREAL~\cite{varol17_surreal} has many subjects, they are synthetic rather than real human subjects. 
In MVPose3D, each subject performs every action four times from two orientations, which is denoted as `$\times$4' in Table~\ref{tab:dataset_cmp}.
As shown in Figure~\ref{fig:dataset_cmp}, MVPose3D offers substantially denser point distributions compared to prior datasets, making it particularly valuable for 3D human pose estimation.
We visualize data samples of MVPose3D dataset in Figure~\ref{fig:dataset_sequence} and Appendix~\ref{sec:supp_dataset}.


\section{Experiments}
\label{sec:exp}

In this section, we first introduce the implementation details of our method. 
We then compare the proposed method with the state-of-the-art MOVIN~\cite{jang2023movin} and SPiKE~\cite{ballester2025spike}. 
For comparison, we use LiDARHuman26M~\cite{li2022lidarcap}, LIPD~\cite{ren2023lidar}, ITOP-Side~\cite{haque2016towards}, and MVPose3D dataset.
In addition, we conduct an ablation study to demonstrate the effectiveness of each proposed component.

\subsection{Implementation details}
\label{subsec:exp_imple}
The proposed model is implemented based on PyTorch~\cite{paszke2019pytorch}. 
We train our model using AdamW optimizer, using a learning rate of $10^{-4}$ and a weight decay of $10^{-4}$.
The batch size is set to 32, and we choose $T = 4$.
To sample constant numbers of points from the point cloud, we employ farthest point sampling (FPS)~\cite{eldar1997farthest} algorithm, selecting $N=256$ points.
For the proposed generative process, we use $I=20$ sampling iterations.
We separately train Point2Pose using two different types of generative process introduced in Sec.~\ref{subsec:method_imu_reg}.
Detailed experimental setup is described in Appendix~\ref{sec:supp_exp_detail}.
In addition, to eliminate noise in the LiDARHuman26M dataset~\cite{li2022lidarcap}, we applied graph-based point cloud clustering (GPC) during training our method, which is explained in the Appendix~\ref{sec:supp_gpc}.

\subsection{Quantitative evaluation}
\label{subsec:exp_quan}

For evaluation, we measure 1) MPJPE for root-relative joint coordinates and 2) Angular Error for joint rotations.
Lower metric value means better performance of the model.
The units of MPJPE and angle error are millimeters and degree, respectively.
\textbf{Bold} and \underline{underlined} numbers indicate the best and second-best performance in each column of the table, respectively.
Unless mentioned otherwise, we use the model predictions as inputs for previous frames for algorithms which requires pose history (Point2Pose and MOVIN~\cite{jang2023movin}).

\paragraph{Comparison with MOVIN.} We choose MOVIN~\cite{jang2023movin} as the direct baseline and compare it with the proposed method using publicly available datasets~\cite{li2022lidarcap, ren2023lidar} and our proposed MVPose3D dataset.
Since the official implementation of MOVIN is unavailable, we reproduce it based on its original paper and attach it to the supplementary material for validation.
We firstly evaluate each algorithm by using ground-truth poses as input for previous frames in both MOVIN and Point2Pose.
Table~\ref{tab:cmp_baselines} indicates that the proposed method consistently outperforms MOVIN.

We also evaluate auto-regressive pose estimation performance -- using model predictions as inputs for previous frames instead of ground-truth.
Note that for inference, we used a smaller number of iterations ($T_1$) than those used during training ($T_2, T_1<T_2$) to mitigate the effect of error accumulation, which is denoted as ``$T=T_2 \rightarrow T=T_1$''.
As shown in Table~\ref{tab:cmp_baselines_ar}, our method performs better, demonstrating robustness against error accumulation.

\begin{table}[t!]
	\centering
	\caption{
		Quantitative comparisons of our method with MOVIN~\cite{jang2023movin} using GT poses as input for previous frames. 
	}
	\vspace{-2mm}
	\setlength\tabcolsep{1.8pt} 
	\scalebox{0.7}{
			\hspace{-2mm}
			\begin{tabular}{l cc cc cc}
				\toprule 
				\multirow{2}{*}{Method} 
				& \multicolumn{2}{c}{LiDARHuman26M~\cite{li2022lidarcap}}
				& \multicolumn{2}{c}{LIPD~\cite{ren2023lidar}}
				& \multicolumn{2}{c}{MVPose3D} \\
				
				\cmidrule(lr){2-3} \cmidrule(lr){4-5} \cmidrule(lr){6-7}  
				& MPJPE 
				& Ang. Err 
				
				& MPJPE 
				& Ang. Err 
				
				& MPJPE 
				& Ang. Err

				\\ 
				\cmidrule(lr){1-7}
				
				{MOVIN~\cite{jang2023movin}}
				&58.56  &25.68
                &59.13 &17.71
                &72.00	&15.16	 \\
				
				\hdashline
				
				{{Point2Pose}$_{\mathrm{DM}}$}
				&\underline{36.23}	&\underline{5.72}
                &\underline{40.52} &\underline{8.72}
                &\textbf{37.02}	&\underline{4.35} \\
				
				{{Point2Pose}$_{\mathrm{OT-CFM}}$}
				&\textbf{35.40}	&\textbf{5.61}
                &\textbf{38.01} &\textbf{8.01}
                & \underline{37.89}	&\textbf{3.91} \\
                
				\bottomrule 
			\end{tabular}
	}
	\vspace{-2mm}
	\label{tab:cmp_baselines}
\end{table}

\begin{table}[t!]
    \centering
    \caption{{Quantitative comparisons of our method with MOVIN~\cite{jang2023movin} using the LIPD~\cite{ren2023lidar} dataset.}}
    \vspace{-2mm}
    \setlength\tabcolsep{1.8pt} 
    \scalebox{0.8}{
    \begin{tabular}{lcc}
    \toprule
    \multirow{2}{*}{Method}  
    & \multicolumn{2}{c}{LIPD~\cite{ren2023lidar}} \\
    \cmidrule(lr){2-3}
    & MPJPE ($\downarrow$)
    & Ang. Err ($\downarrow$) \\ 
    \midrule
    MOVIN~\cite{jang2023movin} & 291.80 & 24.02 \\
    \hdashline
    {{Point2Pose}$_{\mathrm{OT-CFM}}$} ($T=4 \rightarrow T=3$) & \underline{192.54} & 13.80 \\
    {{Point2Pose}$_{\mathrm{OT-CFM}}$} ($T=4 \rightarrow T=2$) & 195.39 & 13.72 \\
    {{Point2Pose}$_{\mathrm{DM}}$} ($T=4 \rightarrow T=3$) & \textbf{189.72} & \textbf{13.48} \\
    {{Point2Pose}$_{\mathrm{DM}}$} ($T=4 \rightarrow T=2$) & 196.73 & \underline{13.67} \\
    \bottomrule
    \end{tabular}
    }
    \vspace{-3mm}
    \label{tab:cmp_baselines_ar}
\end{table}

\paragraph{Comparison with SPiKE.} 

We include SPiKE~\cite{ballester2025spike} as an additional baseline using the ITOP-Side~\cite{haque2016towards} dataset, using its publicly available pretrained checkpoint to evaluate MPJPE. 
Angular Error is not reported since SPiKE does not estimate joint rotations.
As demonstrated in Table~\ref{tab:cmp_baselines_spike}, our method outperforms SPiKE.

\begin{table}[h!]
    \centering
    \caption{{Quantitative comparison of our method with SPiKE~\cite{ballester2025spike} using the ITOP-Side~\cite{haque2016towards} dataset.}}
    \vspace{-2mm}
    \setlength\tabcolsep{1.8pt} 
    \scalebox{0.8}{
    \begin{tabular}{lcc}
    \toprule
    \multirow{2}{*}{Method}  
    & {ITOP-Side~\cite{haque2016towards}} \\
    \cmidrule(lr){2-2}
    & MPJPE ($\downarrow$)\\ 
    \midrule
    SPiKE~\cite{ballester2025spike} & 142.83 \\
    \hdashline
    {{Point2Pose}$_{\mathrm{OT-CFM}}$} ($T=4 \rightarrow T=3$) & {125.01} \\
    {{Point2Pose}$_{\mathrm{OT-CFM}}$} ($T=4 \rightarrow T=2$) & \textbf{124.25}  \\
    \bottomrule
    \end{tabular}
    }
    \vspace{-5mm}
    \label{tab:cmp_baselines_spike}
\end{table}

\paragraph{Evaluation on scenarios lacking initial pose information.}

Notably, our method generalizes well to scenarios without initial pose information. 
By initializing the start pose from Gaussian noise ($\mathcal{N}(\mathbf{0}, \mathbf{I})$), we evaluated on the ITOP-Side~\cite{haque2016towards} dataset.
As shown in Table~\ref{tab:cmp_lack_prior}, performance is comparable to using the initial ground-truth pose, demonstrating Point2Pose's robustness to initial pose uncertainty.

\begin{table}[h!]
    \centering
    \caption{{Quantitative evaluations under scenarios lacking initial pose information using the ITOP-Side~\cite{haque2016towards} dataset.}}
    \vspace{-2mm}
    \setlength\tabcolsep{1.8pt} 
    \scalebox{0.8}{
    \begin{tabular}{lcc}
    \toprule
    \multirow{2}{*}{Method}  
    & {ITOP-Side~\cite{haque2016towards}} \\
    \cmidrule(lr){2-2}
    & MPJPE ($\downarrow$)\\ 
    \midrule
    {{Point2Pose}$_{\mathrm{OT-CFM}}$} (Start from GT, $T=4 \rightarrow T=3$) & \textbf{125.01} \\
    {{Point2Pose}$_{\mathrm{OT-CFM}}$} (Start from noise, $T=4 \rightarrow T=3$) & {129.63}  \\
    \bottomrule
    \end{tabular}
    }
    \vspace{-6mm}
    \label{tab:cmp_lack_prior}
\end{table}

\paragraph{Evaluation on sparse input scenarios.} 
We report quantitative results of the pose estimation under sparse input scenarios using the LIPD~\cite{ren2023lidar} dataset in Table~\ref{tab:cmp_sparse}. 
Here, $N$ denotes the number of points in the point cloud of a single frame used during training and inference.
The results demonstrate the robustness of our method to the density of point clouds.
In addition, the proposed method consistently outperforms MOVIN~\cite{jang2023movin} and maintains strong performance even under sparse input situations.

\begin{table}[h!]
    \centering
    \caption{{Quantitative evaluations under sparse input scenarios using the LIPD~\cite{ren2023lidar} dataset.}}
    \vspace{-2mm}
    \setlength\tabcolsep{1.8pt} 
    \scalebox{0.85}{
    \begin{tabular}{lcc}
    \toprule
    \multirow{2}{*}{Method}  
    & \multicolumn{2}{c}{LIPD~\cite{ren2023lidar}} \\
    \cmidrule(lr){2-3}
    & MPJPE ($\downarrow$)
    & Ang. Err ($\downarrow$) \\ 
    \midrule
    {{Point2Pose}$_{\mathrm{OT-CFM}}$} \\ \quad $T=4 \rightarrow T=3$, $N=256$ & {192.54} & 13.80 \\
      \quad  $T=4 \rightarrow T=2$, $N=128$ & \textbf{183.86} & \textbf{13.80} \\
      \quad  $T=4 \rightarrow T=3$, $N=64$ & 191.36 & 15.24 \\
    \bottomrule
    \end{tabular}
    }
    \vspace{-7mm}
    \label{tab:cmp_sparse}
\end{table}

\paragraph{Evaluation on noisy input scenarios.} 
To evaluate performance in noisy situations, we perturb the point cloud of every frame with Gaussian noise during inference, using two standard deviation values: $\sigma=0.01$ and $\sigma=0.04$. 
We report the quantitative results in Table~\ref{tab:cmp_noisy} using the ITOP-Side~\cite{haque2016towards} dataset.
As expected, performance degrades when input point clouds are perturbed.
However, our method still achieves results comparable to noise-free setting ($\sigma=0.00$), demonstrating the robustness of our method to noisy inputs. 
In particular, for $\sigma=0.01$, our method outperforms SPiKE~\cite{ballester2025spike}, as  discussed in Table~\ref{tab:cmp_baselines_spike}.

\begin{table}[h!]
    \centering
    \caption{{Quantitative evaluations under noisy input scenarios using the ITOP-Side~\cite{haque2016towards} dataset.}}
    \vspace{-2mm}
    \setlength\tabcolsep{1.8pt} 
    \scalebox{0.85}{
    \begin{tabular}{lcc}
    \toprule
    \multirow{2}{*}{Method}  
    & {ITOP-Side~\cite{haque2016towards}} \\
    \cmidrule(lr){2-2}
    & MPJPE ($\downarrow$)\\ 
    \midrule
    {{Point2Pose}$_{\mathrm{OT-CFM}}$} \\ \quad $T=4 \rightarrow T=3$, $\sigma = 0.00$ & \textbf{125.01} \\
    \quad $T=4 \rightarrow T=2$, $\sigma = 0.01$ & 141.59  \\
    \quad $T=4 \rightarrow T=2$, $\sigma = 0.04$ & 156.27  \\
    \bottomrule
    \end{tabular}
    }
    \vspace{-7mm}
    \label{tab:cmp_noisy}
\end{table}

\paragraph{Evaluation on novel sensor setting scenarios.}
To evaluate our method's performance under novel sensor settings, we simulate scenarios in which the sensor-subject distance is larger than during training, reducing the number of points per frame. 
Using a model trained with $N=256$ points, we tested with $N=128$, $64$, and $32$ points.
Results using the LIPD~\cite{ren2023lidar} and ITOP-Side~\cite{haque2016towards} are reported in Table~\ref{tab:cmp_novel_sensor}.
Our method shows comparable performance under novel sensor setting, indicating the generalizability of Point2Pose.

\begin{table}[h!]
    \centering
    \caption{{Quantitative evaluations under a novel sensor setting using the LIPD~\cite{ren2023lidar} and ITOP-Side~\cite{haque2016towards} dataset.}}
    \vspace{-2mm}
    \setlength\tabcolsep{1.8pt} 
    \scalebox{0.75}{
    \begin{tabular}{lccc}
    \toprule
    \multirow{2}{*}{Method}  
    & \multicolumn{2}{c}{LIPD~\cite{ren2023lidar}}
    & \multicolumn{1}{c}{ITOP-Side~\cite{haque2016towards}} \\
    \cmidrule(lr){2-3} \cmidrule(lr){4-4}
    & MPJPE 
    & Ang. Err
    & MPJPE \\ 
    \midrule
    {{Point2Pose}$_{\mathrm{OT-CFM}}$} \\ \quad $T=4 \rightarrow T=3$, $N=256$ & \textbf{192.54} & 13.80 & \textbf{125.01}\\
    \quad $T=4 \rightarrow T=3$, $N=256 \rightarrow N=128$ & 	193.67	& 13.81 &125.32 \\
     \quad $T=4 \rightarrow T=3$, $N=256 \rightarrow N=64$ & 197.76 &	13.60 &	131.19 \\
     \quad $T=4 \rightarrow T=3$, $N=256 \rightarrow N=32$ & 205.76 &	\textbf{13.49} & 145.19
 \\
    \bottomrule
    \end{tabular}
    }
    \vspace{-3mm}
    \label{tab:cmp_novel_sensor}
\end{table}

\subsection{Qualitative results}
\label{subsec:exp_qual}
\begin{figure}[h!]
	\centering
	\includegraphics[width=1.0\linewidth]{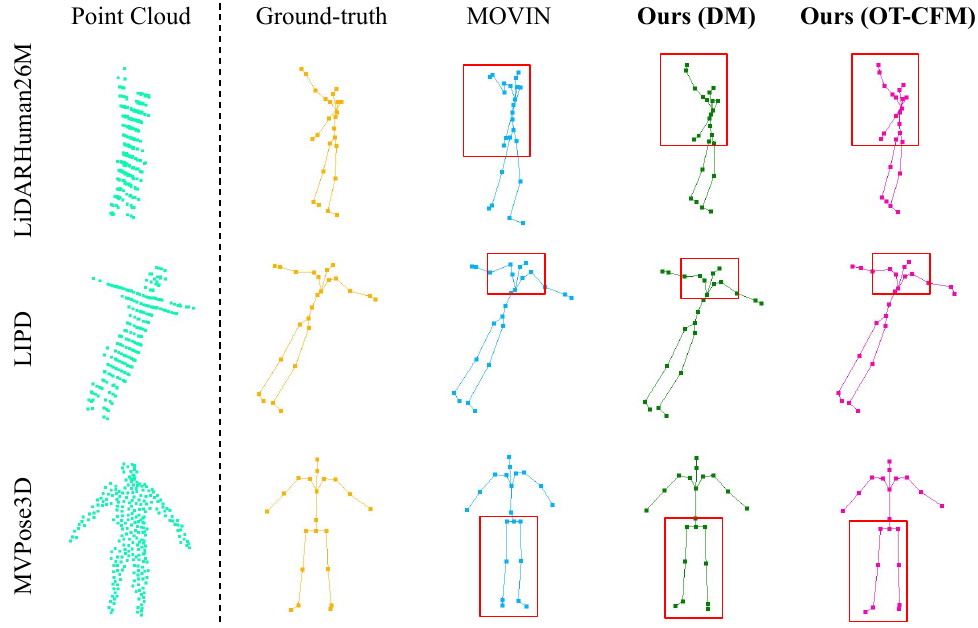}
	\caption{Random results for 3D human pose estimation using MOVIN~\cite{jang2023movin} and the proposed method on LiDARHuman26M~\cite{li2022lidarcap}, LIPD~\cite{ren2023lidar}, and MVPose3D dataset.
    MOVIN fails to accurately predict 3D human pose, with notable errors in upper body (1st row), shoulder joints (2nd-3rd row), and lower body (3rd row).
    In contrast, our method demonstrates better estimation performance.}
    \vspace{-3mm}
\label{fig:qual}
\end{figure}

Figure~\ref{fig:qual} illustrates randomly sampled qualitative results of MOVIN~\cite{jang2023movin} and the proposed method on various datasets.
As visualized, our method accurately estimates joint coordinates, while MOVIN often fails to regress the fine details and plausible distribution of human joints.
We also visualize additional qualitative results using the LiDARHuman26M~\cite{ren2023lidar} and LIPD~\cite{ren2023lidar} in Figure~\ref{fig:teaser} and the Appendix~\ref{sec:supp_qual}.
Note that for both Figure~\ref{fig:teaser} and ~\ref{fig:qual}, we use ground-truth poses as input of  previous frames for every methods.

\subsection{Ablation study}
\label{subsec:exp_ablation}

We show the effectiveness of each proposed component with Point2Pose$_{\mathrm{OT-CFM}}$ model in Table~\ref{tab:ablation}.
Note that we present the additional ablation study on the spatio-temporal point cloud encoder in Appendix~\ref{sec:supp_ablation}.
\vspace{-3mm}
\begin{table}[h!]
	\centering
	\caption{
		Ablation study results for analyzing the significance of each proposed component. 
        Bold and underlined numbers indicates the best and second-best performance in each column, respectively.
        Lower MPJPE value indicates better performance.
	}
	\vspace{-3mm}
	\setlength\tabcolsep{1.8pt} 
	\scalebox{0.78}{
		\begin{tabular}{c l c c}
			\toprule 
			\multirow{2}{*}{Case}
			& \multirow{2}{*}{Description} 
			& {LiDARHuman26M~\cite{li2022lidarcap}}
			& {LIPD~\cite{ren2023lidar}} \\
			
			\cmidrule(lr){3-3} \cmidrule(lr){4-4}
			&
			& MPJPE
			
			& MPJPE

			\\ 
			\cmidrule(lr){1-4}
			
			{(a)}&{w/o global feature}
			&36.17  &{39.47}	\\

			{(b)}&{w/o local feature}
			&37.30	&\underline{38.26}	\\
			
			{(c)}&{w/o temporal attention}
			&\underline{35.79}	&40.00	\\
			
			{(d)}&{w/o spatial attention}
			&36.72	&39.32	\\

			{(e)}&{w/o pose-point feature encoder}
			&101.48	&72.15	\\

			{(f)}&{w/o factorized denoiser}
			&43.63	&54.03	\\

			\hdashline
			
			{} & {Point2Pose$_{\mathrm{OT-CFM}}$ (Ours)}
			&\textbf{35.40}	&\textbf{38.01} \\
			\bottomrule 
		\end{tabular}
	}
	\vspace{-6mm}
	\label{tab:ablation}
\end{table}

\vspace{-2mm}

\paragraph{Spatio-temporal point cloud encoder.}
Case (a)-(d) shows the effects of each components related to point cloud encoder.
Using both local and global features results in powerful representation of the human body.
Also, factorized spatio-temporal attention mechanism makes model to capture continuous geometry features of the human joints, leading to more accurate pose regression.
We use average pooling instead of MHSA for case (c) and (d).

\vspace{-2mm}

\paragraph{Joint-wise pose-point feature encoder.}
We train the proposed model without using joint-wise pose-point feature encoder by substituting $\mathbf{F}_{px} = \mathrm{Duplicate}(\mathbf{F}_p, J)$.
Case (e) shows the importance of joint-wise pose-point feature.

\vspace{-2mm}

\paragraph{Factorized generative pose regressor.}
We jointly denoise $\mathbf{\tilde{J}}_i$ and $\mathbf{\tilde{R}}_i$ instead of using factorized denoising mechanism.
Case (f) shows that the proposed factorized approach is critical to accurately model the pose data distribution.

\vspace{-2mm}

\paragraph{Effects of $T$.}

We show the impact of varying $T$ in Table~\ref{tab:cmp_t} using the LIPD~\cite{ren2023lidar} and ITOP-Side~\cite{haque2016towards} dataset, respectively. 
When $T<4$, the model struggles to accurately estimate poses, emphasizing the importance of pose history information and effectiveness of the proposed joint-wise point-pose feature encoder. 
When $T>4$, MPJPE also increases.
We hypothesize that a larger temporal window ($T$) makes it more difficult for the model to learn consistent temporal dynamics, potentially due to the increased variability in longer sequences.

\begin{table}[h!]
    \centering
    \caption{{Analyzing the effects of $T$ using the LIPD~\cite{ren2023lidar} and ITOP-Side~\cite{haque2016towards} dataset.}}
    \vspace{-2mm}
    \setlength\tabcolsep{1.8pt} 
    \scalebox{0.85}{
        \begin{tabular}{lccc}
        \toprule
        \multirow{2}{*}{Method}  
        & \multicolumn{2}{c}{LIPD~\cite{ren2023lidar}}
        & \multicolumn{1}{c}{ITOP-Side~\cite{haque2016towards}} \\
        \cmidrule(lr){2-3} \cmidrule(lr){4-4}
        & MPJPE  ($\downarrow$)
        & Ang. Err ($\downarrow$)
        & MPJPE  ($\downarrow$) \\ 
        \midrule
        Point2Pose$_{\mathrm{OT-CFM}}$ \\ \quad $T=2 \rightarrow T=2$ & 434.59	& 90.98 & 632.76\\
        \quad $T=3 \rightarrow T=3$ & 451.27	& 86.01 & 507.13 \\
        \quad $T=4 \rightarrow T=3$ & \textbf{192.54}	& \textbf{13.80} & \textbf{125.01}\\
        \quad $T=8 \rightarrow T=3$ & 202.77	& 21.81 & 173.52 \\
        \quad $T=16 \rightarrow T=3$ & 199.87 & 13.97 & 203.28\\
        \bottomrule
        \end{tabular}
	}    
    \vspace{-4mm}
    \label{tab:cmp_t}
\end{table}


\section{Conclusion}
\label{sec:conclusion}

We proposed a novel generative framework for 3D human pose estimation, addressing key challenges in modeling the complex geometry and distribution of human poses. 
Our approach employs a factorized generative pose regressor that iteratively denoises the perturbed data, leveraging encoded point-pose features as conditions to estimate the true distribution of human poses.
In addition, we introduced a large-scale real-world human motion dataset, contributing to the acceleration of human pose estimation research by providing diverse and challenging motion data. 
Extensive experiments on various datasets demonstrate that our method outperforms baseline algorithms in diverse scenarios including sparse scenes, noisy scenes, and scenes without initial pose information.
In conclusion, our method achieves remarkable performance for 3D human pose estimation.

\clearpage
{
    \small
    \bibliographystyle{ieeenat_fullname}
    \bibliography{main}
}

\clearpage
\appendix

\section*{Appendix}
\section{Pseudo-code of Point2Pose}
\label{sec:supp_pseudo_code}
We explain the general training procedure of Point2Pose$_{\mathrm{DM}}$ in Algorithm~\ref{alg:algorithm_train}, and inference process in Algorithm~\ref{alg:algorithm_inference}.


\begin{algorithm}[h!]
	\caption{Training Point2Pose with DM}
	\label{alg:algorithm_train}
	\begin{algorithmic}[1]

		\State \textbf{Inputs:} Sequential point clouds $\mathbf{P}$, Ground-truth pose data $\mathbf{X}^{\mathrm{gt}}$, sampling steps $I$

		\Repeat
        
        \State Compute $\mathbf{F}_{px}$ using Eq.~\eqref{eq:pc_imu_feature}

        \State Sample $\mathbf{\epsilon} \sim \mathcal{N}(0, \mathbf{I})$, $i \sim \mathrm{Uniform}([0, I])$

        \State Calculate $\mathbf{\tilde{X}}_i \gets \mathrm{Perturb}_{\mathrm{DM}}(\mathbf{X}^{\mathrm{gt}},  i)$

        \State Calculate $\mathbf{{X}}^{\mathrm{pred}}_i$ using Eq.~\eqref{eq:denoising_overview}

        \State Calculate $\mathcal{L}$ using Eq.~\eqref{eq:loss_fn}

        \State Optimize $\theta$ with SGD on $\nabla_{\theta} \mathcal{L}$
        \Until{converged}

		\State {\bf Output:} Optimized Point2Pose parameter $\theta^*$
	\end{algorithmic}
\end{algorithm}


\begin{algorithm}[h!]
	\caption{Sampling $\mathbf{X}^{\mathrm{pred}}$ from {Point2Pose} with DM}
	\label{alg:algorithm_inference}
	\begin{algorithmic}[1]

		\State \textbf{Inputs:} Sequential point clouds $\mathbf{P}$, sampling steps $I$, optimized {Point2Pose} parameters $\theta^*$

        \State Compute $\mathbf{F}_{px}$ using Eq.~\eqref{eq:pc_imu_feature}

        \State Sample $\mathbf{\tilde{J}}_I \sim \mathcal{N}(0, \mathbf{I})$, $\mathbf{\tilde{R}}_I \sim \mathcal{N}(0, \mathbf{I})$
        
        \For {$i \gets I, \cdots, 1$}

        \State Calculate $\mathbf{\tilde{J}}_{i-1}$ using Eq.~\eqref{eq:dm_inference_perturb_coord}
        
        \EndFor
        
        \For {$i \gets I, \cdots, 1$}

        \State Calculate $\mathbf{\tilde{R}}_{i-1}$ using Eq.~\eqref{eq:dm_inference_perturb_rotation}
        
        \EndFor

        \State $\mathbf{X}^{\mathrm{pred}} \gets [\mathbf{J}^{\mathrm{pred}}_0, \mathbf{R}^{\mathrm{pred}}_0]$

		\State {\bf Output:} Sampled pose data $\mathbf{X}^{\mathrm{pred}} $
	\end{algorithmic}
\end{algorithm}

\section{Additional ablation study}
\label{sec:supp_ablation}

To show the effectiveness of the spatio-temporal point cloud encoder, we conduct an additional ablation study.
Especially, we set $T=1$ for the point cloud encoder and train  Point2Pose$_{\mathrm{OT-CFM}}$ model using the ITOP-Side~\cite{haque2016towards} dataset. 
Notably, as shown in Table~\ref{tab:supp_ablation}, the performance drops when only the final frame of the point cloud is used, demonstrating the necessity of a spatio-temporal point cloud encoder module.

\begin{table}[h!]
    \centering
    \caption{{Additional ablation study on the temporal encoding of the point cloud encoder using the ITOP-Side~\cite{haque2016towards} dataset.}}
    \setlength\tabcolsep{1.8pt} 
    \scalebox{0.7}{
    \begin{tabular}{lcc}
    \toprule
    \multirow{2}{*}{Method}  
    & {ITOP-Side~\cite{haque2016towards}} \\
    \cmidrule(lr){2-2}
    & MPJPE ($\downarrow$)\\ 
    \midrule
    {{Point2Pose}$_{\mathrm{OT-CFM}}$} ($T=4 \rightarrow T=3$) & \textbf{125.01} \\
    {{Point2Pose}$_{\mathrm{OT-CFM}}$} (only $T=1$ for PC encoder, $T=4 \rightarrow T=3$) & 154.35  \\
    \bottomrule
    \end{tabular}
    }
    \vspace{-6mm}
    \label{tab:supp_ablation}
\end{table}

\section{Additional qualitative results}

\label{sec:supp_qual}

\begin{figure*}[h!]
	\centering
	\includegraphics[width=0.9\linewidth]{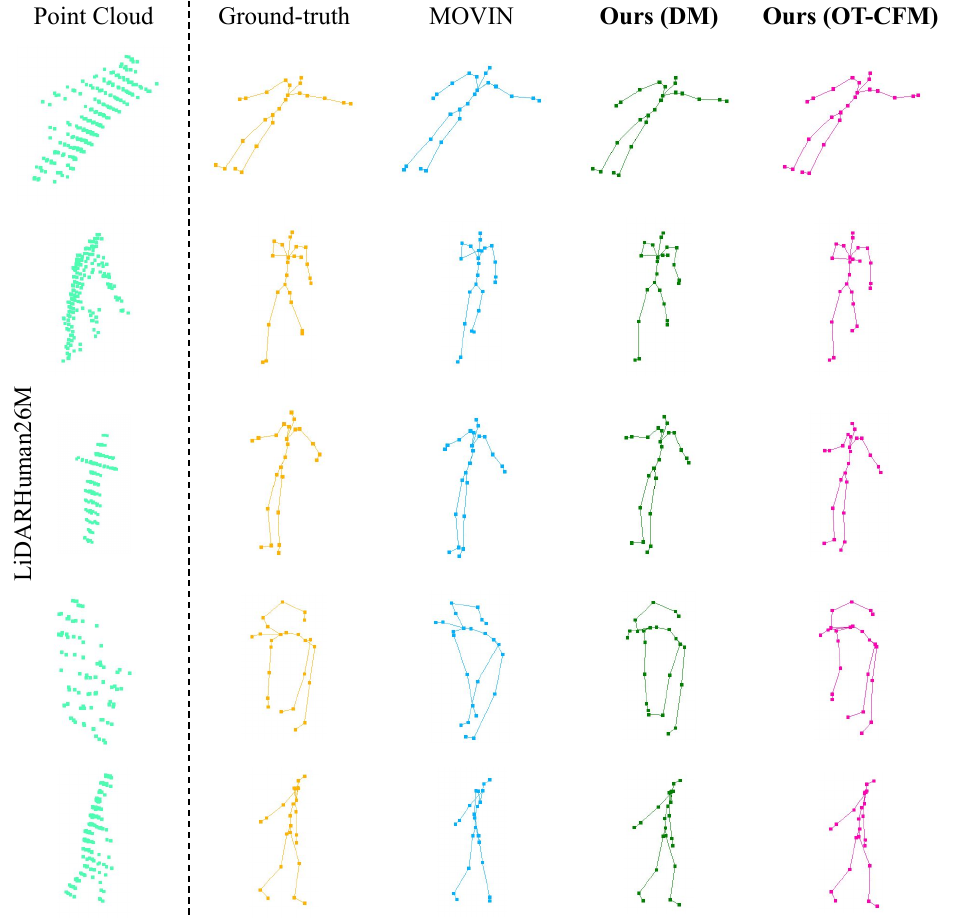}
	\vspace{-0.3cm}
	\caption{Qualitative results of pose estimation using the proposed method and MOVIN~\cite{jang2023movin} on the LiDARHuman26M~\cite{li2022lidarcap} dataset.
    For both Point2Pose and MOVIN, the ground truth  pose is used as input to the model.
    Point2Pose demonstrates more accurate pose estimation compared to MOVIN. }
\label{fig:supple_lidarhuman26m_result}
\end{figure*}

\begin{figure*}[h!]
	\centering
	\includegraphics[width=0.8\linewidth]{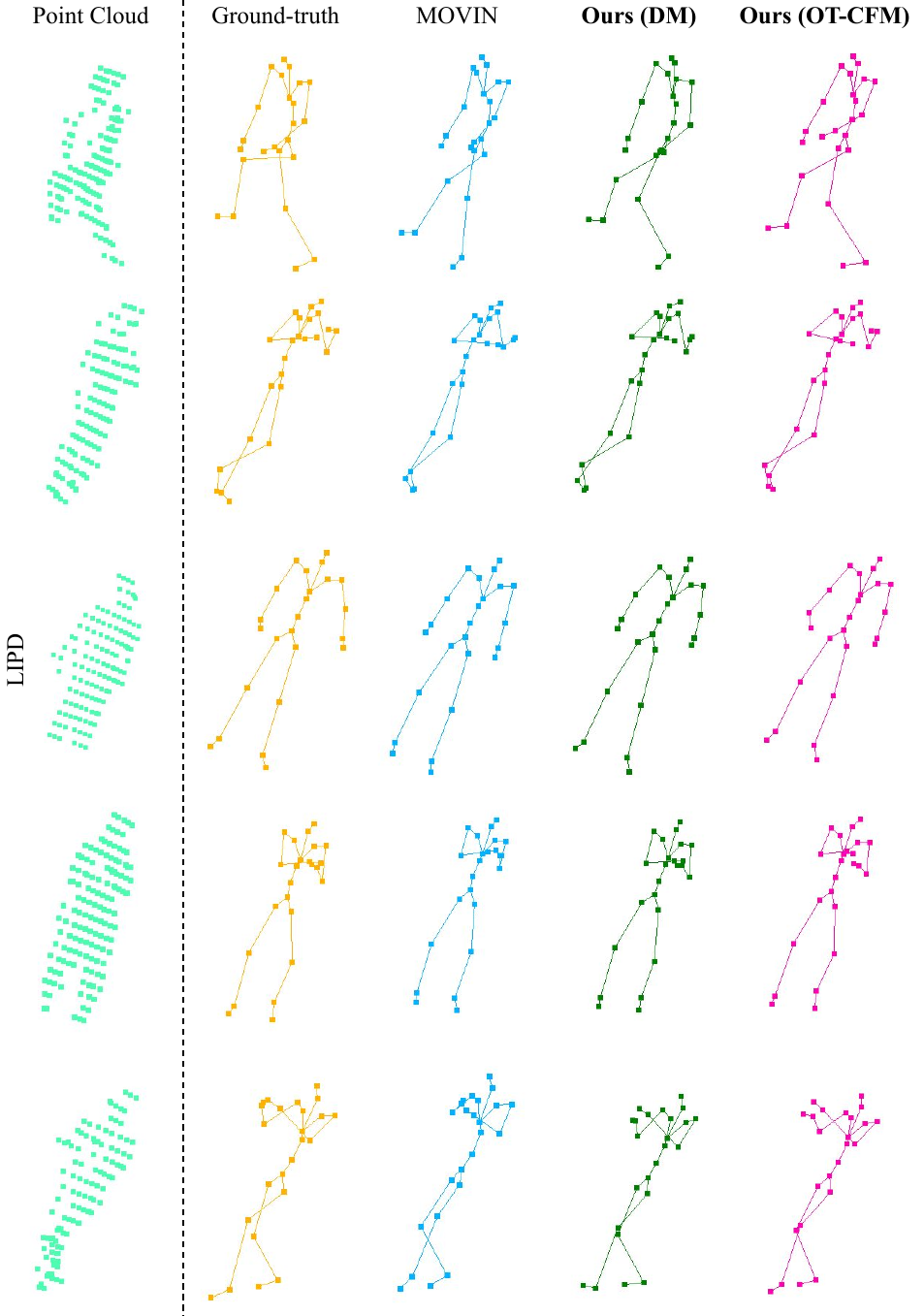}
	\vspace{-0.3cm}
	\caption{Qualitative results of pose estimation using the proposed method and MOVIN~\cite{jang2023movin} on the LIPD~\cite{ren2023lidar} dataset.
    For both Point2Pose and MOVIN, the ground truth  pose is used as input to the model.
    Point2Pose demonstrates more accurate pose estimation compared to MOVIN. }
\label{fig:supple_lipd_result}
\end{figure*}

We visualize additional results of our method and MOVIN~\cite{jang2023movin} on the LiDARHuman26M~\cite{ren2023lidar} and LIPD~\cite{ren2023lidar} dataset in Figure~\ref{fig:supple_lidarhuman26m_result} and~\ref{fig:supple_lipd_result}, respectively. 
As shown, our method achieves outstanding performance compared to MOVIN.

\section{Graph-based point cloud clustering}
In this section, we discuss about the \textit{graph-based point cloud clustering (GPC)} mentioned in Section~\ref{subsec:exp_imple}, which is a novel strategy used for eliminating noise in the LiDARHuman26M~\cite{li2022lidarcap} dataset.

\label{sec:supp_gpc}

Point clouds which captures humans in a real world often contains noisy points, due to the technical limitations of capture devices.
To address the issue, we propose a novel point cloud pre-processing method which effectively eliminates noisy points using graph-based clustering.
We firstly generate 2D KD-tree $\mathcal{D}$ by projecting input 3D point clouds into $xy$-plane, and extract $n_C$ 2D point clusters $\{ \mathcal{C}_i \}_{i=0}^{n_C}$ with BFS.
Then, to eliminate noisy points, we modify each clusters by
\begin{align}
    {\mathcal{\tilde{C}}_i} &= \{x_j | x_j \in \mathcal{C}_i, |z(x_j) - \mu(\mathcal{C}_i)| < \lambda \cdot \sigma(\mathcal{C}_i)\}, \nonumber \\
    \mathcal{C}_i &= {\mathcal{\tilde{C}}_i} \text{ if }|{\mathcal{\tilde{C}}_i}|\geq n_c \text{ else }\phi 
\end{align}
where $\mu(\mathcal{C}_i)$ and $\sigma(\mathcal{C}_i)$ denotes the mean and standard deviation of depth of points consisting $\mathcal{C}_i$, $z(\cdot)$ denotes depth of given point, and $\lambda$ and $n_c$ is hyperparameter.
Next we select the largest cluster $\mathcal{C}_{i*}$, where $i^* = argmax_i |\mathcal{{C}}_i|$.
Based on $\mathcal{C}_{i*}$, \textit{reachable clusters} $\mathcal{C}_{\mathrm{reach}}$ are calculated by
\begin{align}
    \mathcal{C}_{\mathrm{reach}}= \{\mathcal{{C}}_i | 
    & ^\exists  x_j \in \mathcal{C}_{i*},  x_k \in \mathcal{{C}}_i \text{ s.t. } x_k \in \mathrm{kNN}(x_j), \nonumber \\
    & M_1 < |z(x_k)|<M_2,\;  \; i \neq i^*
    \},
\end{align}
where $M_1$ and $M_2$ are threshold derived by the depth values of points of $\mathcal{C}_{i^*}$, and $\mathrm{kNN}(\cdot)$ returns k-nearest neighbors of given point.
We finally construct clustered points as follows:
\begin{equation}
    \mathcal{C}_{\mathrm{output}} = \{x_i | x_i \in \{ \mathcal{C}_{i^*} \} \cup \mathcal{C}_{\mathrm{reach}} \}.
\end{equation}

\paragraph{Effectiveness of GPC.}
Using LiDARHuman26M dataset, we measured chamfer distance (CD) between human mesh $\mathbf{m}$ reconstructed with SMPL~\cite{SMPL:2015} and point cloud $\mathbf{p}$, which is calculated by
\begin{equation}
	\frac{1}{|\mathbf{p}|} \sum_{x_1 \in \mathbf{p}} \min_{x_2 \in \mathbf{m}} \|x_1 - x_2 \|^2_2 + \frac{1}{|\mathbf{m}|} \sum_{x_1 \in \mathbf{m }} \min_{x_2 \in \mathbf{p}} \|x_1 - x_2 \|^2_2.
\end{equation}
Notably, chamfer distance is decreased from 36.55 to 34.70 by utilizing Graph-based point cloud clustering, which demonstrates its effectiveness on noisy points elimination.

\section{Experimental details}
\label{sec:supp_exp_detail}
Since all of the LiDARHuman26M~\cite{li2022lidarcap}, LIPD~\cite{ren2023lidar} and MVPose3D datasets do not provide joint velocity and acceleration data, we use joint coordinates and rotations to train both Point2Pose and MOVIN.
To train Point2Pose on the LiDARHuman26M and LIPD dataset, we used 30 epochs, halving the learning rate after 20 epochs.
In case of training the proposed method on MVPose3D dataset, we use 20 epochs.
MOVIN is trained for 120 epochs on LiDARHuman26M and LIPD, and for 60 epochs on MVPose3D dataset, with a learning rate of $10^{-4}$.
We do not utilize learning rate scheduling when training MOVIN.
Note that we reproduce MOVIN~\cite{jang2023movin} due to the inaccessibility of its official implementation. For validation, we include our implementation of MOVIN in the supplementary material.
We use single-view point clouds from the MVPose3D for all experiments.

\section{Details on the proposed dataset: MVPose3D}
\label{sec:supp_dataset}

\subsection{Dataset visualization}

\begin{figure*}[ht]
	\centering
	\includegraphics[width=0.8\linewidth]{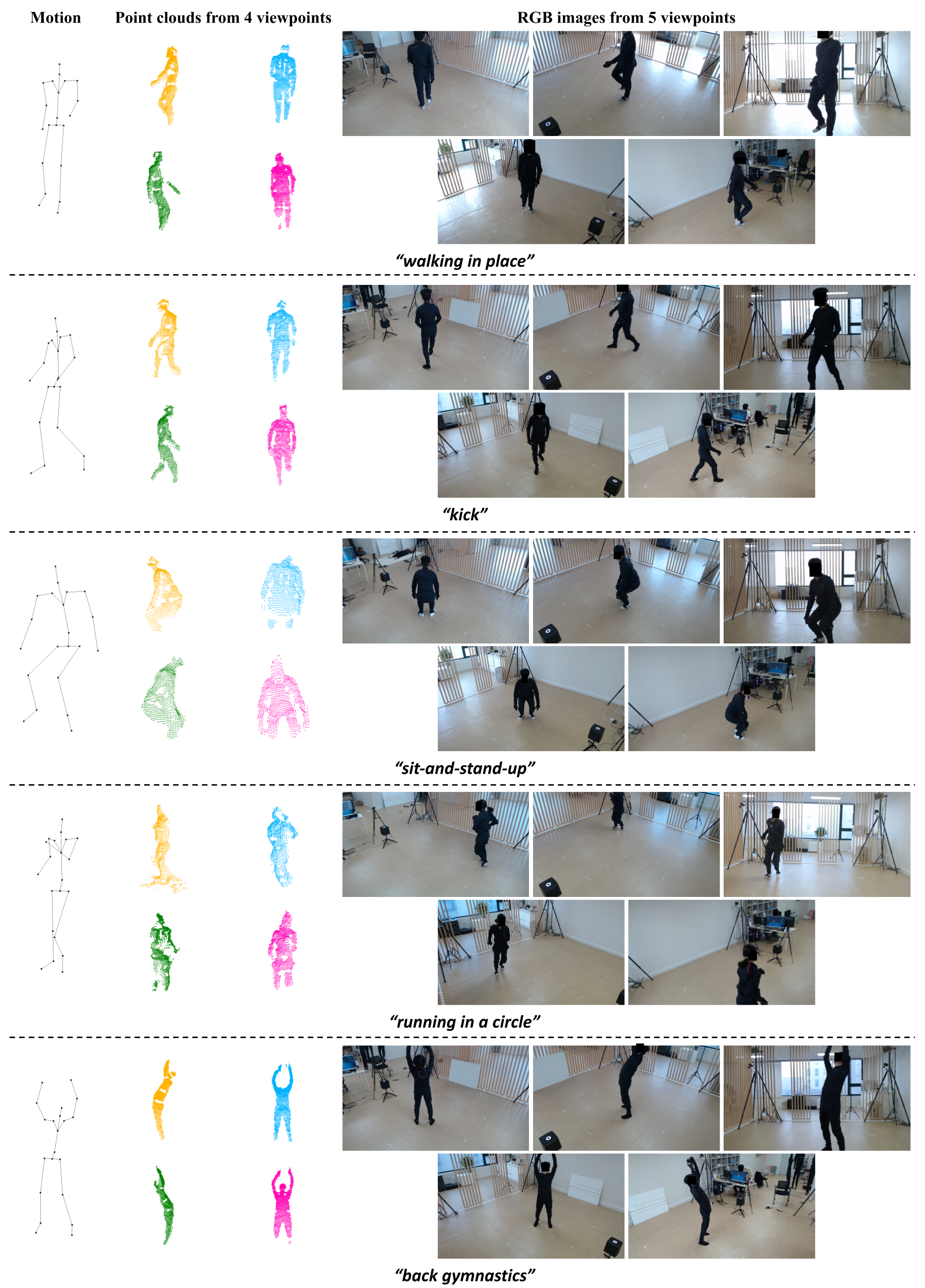}
	\caption{Samples from the proposed dataset.
    The MVPose3D dataset includes a diverse range of human motions.
    Each sample consists of four point clouds, five RGB images, 3D joint coordinates, and joint rotation data.
    We mask each participant's face for visualization. }
\label{fig:supple_dataset_1}
\end{figure*}

\begin{figure*}[ht]
	\centering
	\includegraphics[width=0.85\linewidth]{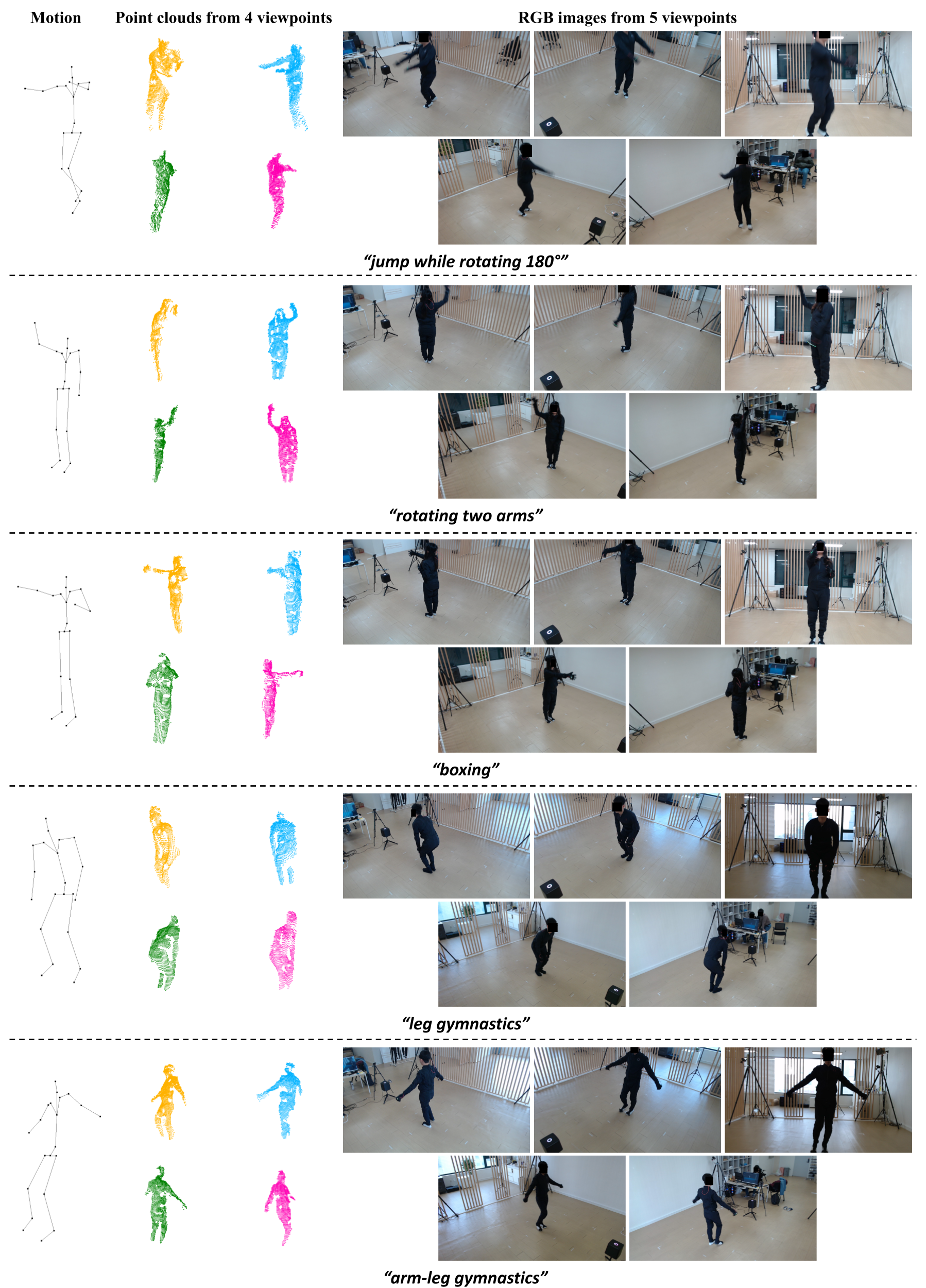}
	\caption{Samples from the proposed dataset.
    The MVPose3D dataset includes a diverse range of human motions.
    Each sample consists of four point clouds, five RGB images, 3D joint coordinates, and joint rotation data.
    We mask each participant's face for visualization.}
\label{fig:supple_dataset_2}
\end{figure*}

We visualize samples from the MVPose3D dataset in Figure~\ref{fig:supple_dataset_1},~\ref{fig:supple_dataset_2}. Each sample consists of four point clouds, five RGB images captured from different viewpoints, and motion data including joint coordinates and rotations. 
As shown in Figure, our dataset involves diverse human motions and dense point clouds.
Note that we mask each participant's identity only for visualization.

\subsection{Data preprocessing}

To segment point cloud corresponding to the human body, we first eliminate the background using a reference depth map, then apply DFS~\cite{tarjan1972depth} for initial segmentation.
Next, we convert the segmented depth map into a point cloud, and multiply the transform matrix obtained through extrinsic calibration.
We further refine the segmented point clouds using DBSCAN~\cite{ester1996density} and Statistical Outlier Removal~\cite{rusu2010semantic}.
The overall process is summarized in Algorithm~\ref{alg:subject_extraction_concise_hs}. 


\begin{algorithm}[h]
\caption{Depth-based human body segmentation}
\label{alg:subject_extraction_concise_hs}
\begin{algorithmic}[1]

    \State \textbf{Input:} Captured depth map $\mathbf{D}\in\mathbb{R}^{M\times N}$, reference map ${\mathbf{D}_{\mathrm{ref}}}\in\mathbb{R}^{M\times N}$, iToF extrinsic parameters $\mathbf{T}_{\mathrm{ext}}$, hyperparameter $\tau$
    
    \For {$u \gets 1, \cdots, M$}
        
        \For {$v \gets 1, \cdots, N$}
            
            \If{$\bigl|\mathbf{D}(u,v) - \mathbf{D}_{\mathrm{ref}}(u,v)\bigr| < \tau$}
                \State  $\mathbf{D}(u,v) \leftarrow 0$
            \EndIf
        
        \EndFor
        
    \EndFor
    \Comment{Background elimination}
    
    \State $\mathbf{C} \leftarrow  \mathrm{Largest}\_\mathrm{connected}\_\mathrm{component}(\mathbf{D})$

    \State $\mathbf{P} \leftarrow  \mathrm{Depth}\_\mathrm{to}\_\mathrm{point}\_\mathrm{cloud}(\mathbf{C})$
    
    \State $\mathbf{P}_{\mathrm{transformed}} = \bigl\{\mathbf{T}_{\mathrm{ext}}\,\bigl(x,y,z,1\bigr)^\top
  \;\Big|\, (x,y,z)\in \mathbf{P} \bigr\}$

    \State $\mathbf{P}_{\mathrm{DB}} \leftarrow \mathrm{DBSCAN}\bigl(\mathbf{P}_{\mathrm{transformed}}\bigr)$
    
    \State $\widetilde{\mathbf{P}} \leftarrow \mathrm{SOR}\bigl(\mathbf{P}_{\mathrm{DB}} \bigr)$
    \Comment{Human body segmentation}
    
    \State \textbf{Output:} Clustered human body point cloud
    \(
      \widetilde{\mathbf{P}}
    \)
    
\end{algorithmic}
\end{algorithm}

\subsection{Details on human resources}

\paragraph{Motion capture from subjects.}

MVPose3D contains 18 distinct actions performed by each subject, covering a wide range of human motions.
Table~\ref{tab:actions} summarizes the details of captured motions.


\begin{table}[ht]
  \centering
  \caption{Types of human actions included in MVPose3D dataset.}
  \hspace{-4mm}
  \setlength\tabcolsep{1.8pt}
  \scalebox{0.95}{
    \begin{tabular}{l l}
        \toprule
        \textbf{Static} & \textbf{Locomotive} \\
        \midrule   
        T-pose, A-pose, Attention & leg, side gymnastics \\
        walking in place, throwing & back, jumping gymnastics \\
        Upper-body T-pose and rotate & double arm-leg gymnastics \\
        kick, boxing & running in a straight line \\
        sit-to-stand & running in a circle \\
        Jump while rotating 180° & Boxing while running \\
        \bottomrule 
    \end{tabular}
  }
  \label{tab:actions}
\end{table}

\paragraph{Potential risks related to study participants.}

There are no potential risks to participants from the data acquisition experiment itself, since it does not involve any equipment or elements that are harmful to humans. 
This information has been informed to the participants in advance.

\section{Computational overhead}
\label{sec:supp_computation}

Using a single NVIDIA GeForce RTX 3090 GPU, our method takes 1.62 hours and requires 19.46 GB of GPU memory to train for 30 epochs with a batch size of 32 on the LIPD~\cite{ren2023lidar} dataset.

\section{Social impacts}
\label{sec:supp_social}

This work can be widely applied across various fields that require precise 3D human pose estimation, such as virtual reality (VR), augmented reality (AR), digital healthcare, and online gaming.
However, due to the limitations of the generative model (DM and OT-CFM) used, it may produce inaccurate or harmful results, which may potentially pose a risk to users.

\section{Limitations}
Due to the limitations of generative models, our method can produce implausible results for extremely abnormal poses. 
In addition, the pose estimation error may increase when dealing with very sparse point cloud data.


\clearpage

\end{document}